\documentclass[sn-mathphys-num]{sn-jnl}

\usepackage{graphicx}%
\usepackage{multirow}%
\usepackage{amsmath,amssymb,amsfonts}%
\usepackage{amsthm}%
\usepackage{mathrsfs}%
\usepackage[title]{appendix}%
\usepackage{xcolor}%
\usepackage{textcomp}%
\usepackage{manyfoot}%
\usepackage{booktabs}%
\usepackage{algorithm}%
\usepackage{algorithmicx}%
\usepackage{algpseudocode}%
\usepackage{listings}%
\usepackage{amsmath}
\usepackage{amssymb}
\usepackage{graphicx}
\usepackage{wasysym}
\usepackage{tabularx}
\usepackage{color, xcolor}

\theoremstyle{thmstyleone}%
\theoremstyle{thmstyletwo}%

\theoremstyle{thmstylethree}%

\begin{document}

\title[Article Title]{X Modality Assisting RGBT Object Tracking}


\author[1]{\fnm{Zhaisheng} \sur{Ding}}\email{dzs@jsnu.edu.cn}

\author*[2]{\fnm{Haiyan} \sur{Li}}\email{leehy@ynu.edu.cn}

\author[3]{\fnm{Ruichao} \sur{Hou}}\email{rc$\_$hou@smail.nju.edu.cn}

\author[2]{\fnm{Yanyu} \sur{Liu}}\email{yanyul054@gmail.com}

\author[2]{\fnm{Shidong} \sur{Xie}}\email{shidongxie@mail.ynu.edu.cn}

\affil*[1]{\orgdiv{School of Electrical Engineering and Automation}, \orgname{Jiangsu Normal University}, \orgaddress{\city{Xuzhou}, \postcode{221116} \country{China}}}

\affil[2]{\orgdiv{School of information}, \orgname{Yunnan University}, \orgaddress{ \city{Kunming}, \postcode{650504}, \country{China}}}

\affil[3]{\orgdiv{State Key Laboratory for Novel Software Technology}, \orgname{Nanjing University}, \orgaddress{\city{Nanjing}, \postcode{210023}, \country{China}}}


\abstract{Developing robust multi-modal feature representations is crucial for enhancing object tracking performance. \textcolor{red}{In pursuit of this objective, a novel X Modality Assisting Network (X-Net) is introduced,} which explores the impact of the fusion paradigm by decoupling visual object tracking into three distinct levels, thereby facilitating subsequent processing. Initially, to overcome the challenges associated with feature learning due to significant discrepancies between RGB and thermal modalities, \textcolor{red}{a plug-and-play pixel-level generation module (PGM) based on knowledge distillation learning is proposed.} This module effectively generates the X modality, bridging the gap between the two patterns while minimizing noise interference. Subsequently, to optimize sample feature representation and promote cross-modal interactions, \textcolor{red}{a feature-level interaction module (FIM) is introduced,} integrating a mixed feature interaction transformer and a spatial-dimensional feature translation strategy. Finally, to address random drifting caused by missing instance features, \textcolor{red}{a flexible online optimization strategy called the decision-level refinement module (DRM) is proposed,} which incorporates optical flow and refinement mechanisms. The efficacy of X-Net is validated through experiments on three benchmarks, demonstrating its superiority over state-of-the-art trackers. \textcolor{red}{Notably, X-Net achieves performance gains of 0.47$\%$/1.2$\%$ in the average of precise rate and success rate, respectively.}  Additionally, the research content, data, and code are pledged to be made publicly accessible at https://github.com/DZSYUNNAN/XNet.}





\maketitle

\section{Introduction}\label{sec1}

Object tracking endeavors to accurately predict a bounding box by harnessing the wealth of information from dual modalities. This approach leverages the complementary cues provided by both RGB and thermal images, enabling uninterrupted operation across various lighting conditions\cite{1}. Benefiting from the leapfrog development of deep learning and thermal imaging techniques, extensive remarkable multi-sensor object tracking methods have been proposed, which greatly improved the precision and success of object tracking, while also bolstering the application of multi-sensor fusion in various practical domains such as autonomous driving, military reconnaissance, and intelligent security systems\cite{2,3}.

RGB images are adept at capturing informative color features and rich textures but are sensitive to environmental illumination conditions. Conversely, thermal images exhibit high stability and are impervious to variations in lighting. They demonstrate strong resilience to challenging environmental factors, such as haze, rain, and snow, but lack the fine details of objects. The amalgamation of complementary features from RGB and thermal images can significantly enhance the classification accuracy of challenging samples, thereby improving tracking accuracy\cite{4}. Therefore, the effective extraction and utilization of this complementary information are of paramount importance in RGBT tracking\cite{5}.

Discrimination-based RGBT trackers have garnered considerable interest due to their exceptional tracking accuracy. For example, Li et al.\cite{12} introduced a discriminative multi-adapter network (MANet) for RGBT tracking. This network effectively exploits the complementary features between modalities and instance perception information by employing multiple convolutional layers with varying kernel sizes. MANet has demonstrated a pronounced ability to extract shared features; however, the extensive use of large convolutional kernels for feature extraction can compromise network efficiency. Tu et al.\cite{13}, on the other hand, proposed a multi-modal multi-margin metric learning (${{\text{M}}^{\text{5}}}{\text{L}}$) tracker to enhance the exploitation of structural information from hard samples and to facilitate a quality-aware fusion of complementary features. Despite these efforts, the ${{\text{M}}^{\text{5}}}{\text{L}}$ method exhibits less-than-satisfactory performance in terms of tracking precision and success rate, falling short of state-of-the-art (SOTA) standards. These mentioned approaches aim to overcome the challenge of mining and utilizing correlation cues among different modalities while also enhancing the anti-interference capabilities of trackers. Nonetheless, there remains significant room for further refinement and improvement.

\begin{figure*}[htbp]
    \centering
    \includegraphics[width=0.98\linewidth]{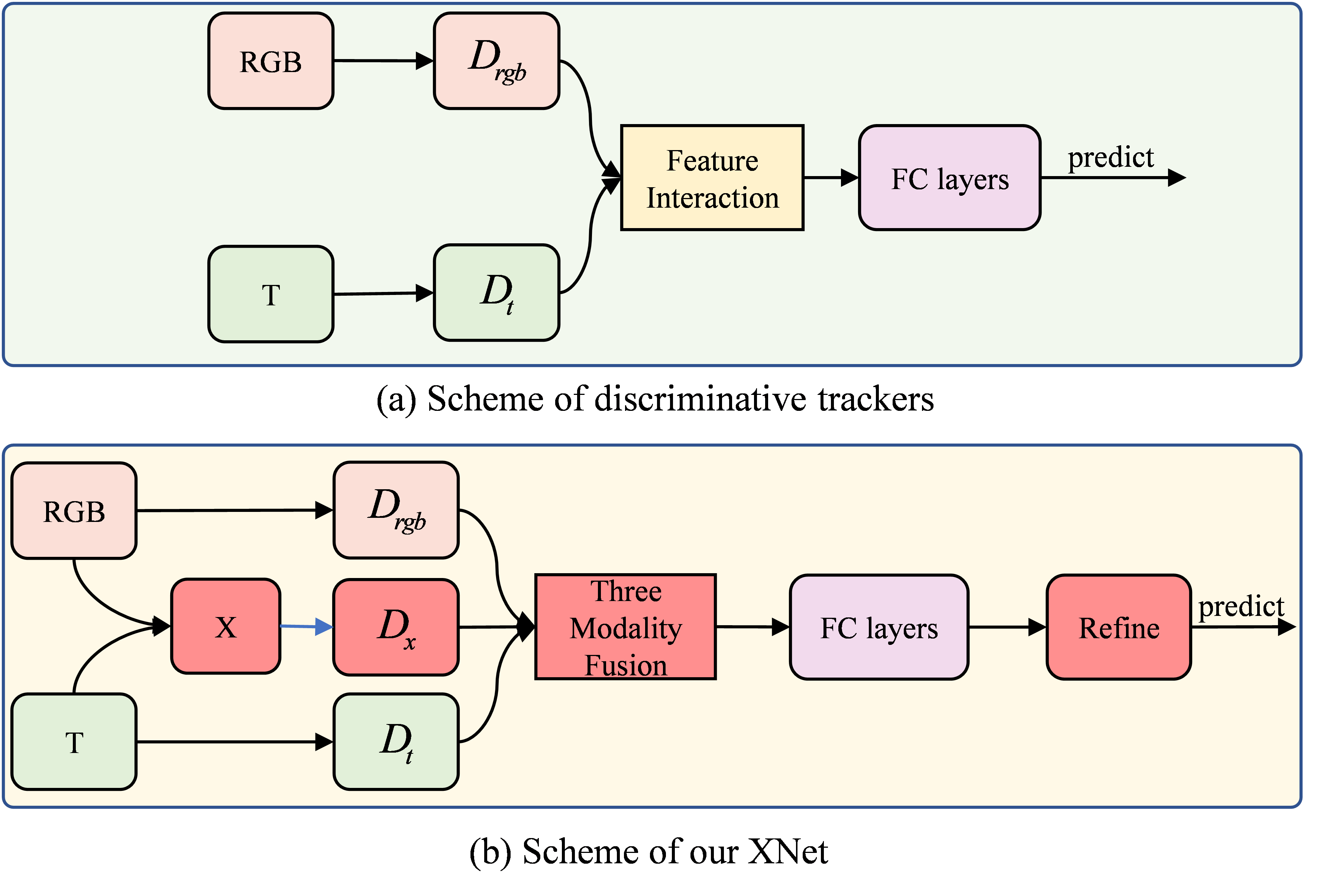}
    \caption{Compared with existing RT-MDNet-based trackers. ${D_{rgb}},{D_t},{D_x}$ denote the deep features of RGB, Thermal and X modality.}
    \label{fig1}
\end{figure*}

In summary, contemporary RGBT tracking systems continue to face several pivotal challenges. Firstly, it remains difficult to ascertain the relative importance of target features across both RGB and infrared thermal imagery. Secondly, efficiently utilizing the inter-modal correlations among diverse feature types is challenging. Finally, existing tracking methods lack an effective mechanism for self-correction. To address these challenges, \textcolor{red}{a novel multi-modal assisting network is proposed, termed as X-Net, which exploits the inter-dependencies between modalities.} X-Net is designed to enhance tracking efficacy through three critical enhancements: pixel-level synthesis, feature-level fusion, and decision-level optimization. As shown in Fig. 1, \textcolor{red}{the proposed network} employs a deep architecture to extract rich features from RGB and thermal images, which are subsequently integrated for target position prediction. The initial phase of X-Net involves synthesizing pixel-level RGB and thermal images to create an X modality that amalgamates complementary features from the source images. Following feature extraction and fusion, the predictions post-classification are refined to reinforce tracking accuracy.

This paper presents the main contributions in a four-fold manner.
\begin{enumerate}
\item{\textcolor{red}{A novel X modality assisting network (X-Net) is proposed} to enhance tracking performance via three pivotal enhancements: pixel-level generation, feature-level interaction and decision-level refinement.}
\item{\textcolor{red}{The lightweight pixel-level generation module (PGM) is integrated} within X-Net. PGM employs a self-knowledge distillation mechanism to generate pixel-level feature aggregation maps, effectively harnessing the capabilities of a high-performance image fusion network to capture modality-shared features.}
\item{The feature interaction module (FIM) is proposed in conjunction with the spatial-dimensional feature translation strategy (SFTS) and the mixed feature interaction transformer (MFIT) to address variation scaling and explore cross-modal global correlations.}
\item{The decision-level refinement module (DRM) is designed, which combines optical flow and refinement mechanisms to optimize tracking results by the flexible tracking strategy. Comprehensive experiments conducted on three RGBT benchmarks demonstrate that the proposed X-Net outperforms existing state-of-the-art trackers.}
\end{enumerate}

\section{Related Work}\label{sec2}

The domain of RGBT tracking has seen substantial progress in recent years, with the development and availability of RGBT tracking datasets serving as a significant milestone. Numerous researchers have endeavored to develop deep learning-based RGBT tracking algorithms aimed at achieving high performance, which are reviewed from two perspectives as follows.

\subsection{Deep Learning-based RGBT Tracking Methods}
RGBT tracking methodologies can be categorized into two principal types based on their tracking mechanisms: generative trackers and discriminative trackers. Generative RGBT trackers rely on a probabilistic model to encapsulate the correlation and joint distribution between thermal infrared and RGB pixels to facilitate object tracking. Trackers based on Siamese networks have gained popularity in the realm of RGBT visual tracking due to their high computational efficiency. For example, SiamFT\cite{14} employed the SiamFC\cite{15} architecture as a foundation for feature extraction across multiple modalities, subsequently merging the features through a manually designed modality weight allocation strategy. Building upon this, DSiamMFT\cite{16} effectively harnesses multi-level semantic features, achieving enhanced accuracy compared to SiamFT. DuSiamRT\cite{17} integrated a channel attention module to derive fusion features from template images. SiamCDA, based on the advanced tracker SiamRPN++, improved the discriminative ability of deep features and tracker robustness by mitigating unimodal discrepancies. SiamCSR\cite{18} combined the channel-spatial attention mechanism with the SiamRPN++ network to perform RGBT tracking. Typically, Siamese network-based trackers exhibit high computational efficiency, fully satisfying the demands of real-world applications. However, their efficacy in handling heterogeneous RGBT data is limited, which constrains their tracking capabilities.

In contrast to generative RGBT tracking approaches, discriminative RGBT trackers integrate the complementary information from visible and infrared imagery to construct a discriminant model for target tracking. The majority of discriminative trackers incorporate an attention mechanism to bolster the characterization of the target, effectively distinguishing it from the background. mfDiMP\cite{19}, for instance, introduced a sophisticated target prediction network, based on DiMP\cite{20}, and utilized discriminant loss for end-to-end training. MANet merged the attention mechanism with an expanded version of MDNet\cite{21}, transforming it into a multi-channel network to extract common and intrinsic features across modalities. MANet++\cite{22} enhanced the features within each modality prior to fusion, thereby surpassing the performance of MANet. MIRNet\cite{23} combined self-attention and cross-attention modules to deliver robust RGBT tracking. Retaining the discriminative characteristics of modal features aids trackers in differentiating the target region from the background—a strategy employed by numerous trackers. For example, the ${{\text{M}}^{\text{5}}}{\text{L}}$ tracker introduced an attention-based fusion module that effectively integrated quality perception across source images, thereby enhancing the accuracy of object tracking. DMCNet\cite{24} proposed a novel dual-gated mutual conditional network to maximize the discriminant information of multi-modalities while repressing noise influence. Both ADRNet\cite{25} and APFNet\cite{26} concentrate on extracting salient features from different image types and blending them to address various attribute challenges. AGMINet\cite{27} presented an asymmetric global and local mutual integration network for the exploitation of heterogeneous features.

In the realm of RGBT object tracking, discriminative trackers tend to have longer processing times compared to generative trackers. Nevertheless, generative trackers capitalize on the feature interactions between different modalities, yielding more robust and precise characteristics. The discriminative trackers face the critical tasks of efficiently extracting intrinsic and shared features from multi-modal images, adeptly combining and enhancing salient features, and employing appropriate tracking strategies. These challenges underscore the need for an advanced architecture to improve tracking efficacy.

\subsection{RGBT information fusion based on knowledge distillation}
The effective fusion of RGB and thermal information is a pivotal aspect of RGBT tracking \cite{53}. Numerous fusion techniques have been proposed, including element-wise summation, concatenation, and content-dependency weighting-based strategies. However, a majority of these fusion methods fail to consider the feature discrepancies between input RGB and thermal images, which, in fact, have distinct imaging mechanisms and exhibit considerable differences in their feature sets \cite{59}. The direct fusion of weighted unimodal RGB and thermal features can diminish the discriminability of the resultant fused features and, consequently, degrade tracking performance. To exploit cross-modal features between RGB and thermal images fully, various RGBT information fusion methods such as Swinfusion\cite{54}, SeAFusion\cite{55}, and CMFA$\_$Net\cite{56} have been proposed, leveraging different deep network architectures to achieve commendable results. These methods capitalize on the complementary nature of infrared and RGB images and utilize carefully crafted feature fusion modules to integrate these characteristics.

Furthermore, delving into the strategies employed by high-performing RGBT fusion networks could yield valuable insights into feature interaction methodologies. Many algorithms leverage knowledge distillation techniques, borrowing advanced fusion strategies from existing models, to inform the development of their own RGBT information fusion processes. For instance, the HKDnet\cite{57} employs a heterogeneous knowledge distillation framework to enable the concurrent fusion and super-resolution of thermal and visible images. This framework includes a high-resolution image fusion network, referred to as the "teacher" network, along with a low-resolution fusion and super-resolution network termed the "student" network. The role of the teacher network is to fuse high-resolution input images and to steer the student network in acquiring the competencies of fusion and super-resolution. This approach yields notable visual enhancements while precisely preserving the natural textures. The CMD\cite{58} tracker incorporates a multi-path selection distillation module that directs a simple fusion module to learn refined multi-modal information from a meticulously designed fusion mechanism. A potential downturn in the performance of RGBT trackers can be attributed to an overemphasis on the expression capabilities of their features.

\textcolor{red}{\subsection{Refinement Mechanism for RGBT Tracking}}
\textcolor{red}{The calculation of the bounding box location is crucial to the accuracy of object tracking methods. Existing approaches typically predict this value through regression; however, such methods suffer from the issue of cumulative failure. In recent years, many methods have adopted refinement strategies to optimize the bounding box prediction, thereby achieving higher tracking accuracy and robustness. Lu et al. designed a refinement strategy based on optical flow calculation to optimize the bounding box location\cite{lu2022duality}. Although the aforementioned tracker achieves more precise performance, the computational complexity of calculating the optical flow field remains a significant challenge. Therefore, Hou et al. proposed a well-refined location prediction module to  estimate the precise scale in the online tracking\cite{23}. Zhao et al. proposed a redetection multimodal fusion network for RGBT tracking, named RMFNet, which provides an adjustment strategy for refining the coarse location boundary box of the target\cite{Zhao_RMFNet}. The adjustment strategy via the refinement network significantly improves the quality of boundary boxes generated by the basic tracker. Tang et al. developed the refinement network to  refine the features of both thermal and RGB light images, which obtains a more precise predicted bounding box\cite{Tang_MFRNet}. Refinement mechanisms have significantly enhanced the performance of trackers; however, effectively leveraging optical flow methods and refinement networks remains an area requiring further investigation.}

\section{Methodology}\label{sec3}

\subsection{Overview}
\begin{figure*}[htbp]
    \centering
    \includegraphics[width=1.0\linewidth]{2.png}
    \caption{Overview of the proposed X-Net framework.}
    \label{fig2}
\end{figure*}
Fig. 2 presents an overview of the X-Net pipeline. Specifically, the backbone of the proposed architecture consists of a customized lightweight VGG-M network equipped with dilated convolutions. \textcolor{red}{The initial three dilated convolution layers are adopted} to extract deep features from the RGB, thermal, and X modalities, as produced by the plug-and-play pixel-level generation module (PGM). Subsequently, \textcolor{red}{a feature-level interaction module (FIM) is designed} by combining a spatial-dimensional feature translation strategy (SFTS) and a mixed feature interaction transformer (MFIT) to obtain an optimal feature expression. Ultimately, \textcolor{red}{a decision-level refinement module (DRM) is proposed} to determine the re-tracking strategy according to the confidence score and motion offset.

\subsection{X Modality Assisting Network}
\textbf{Pixel-level generation module (PGM).} Inspired by the success of infrared and visible image fusion techniques, \textcolor{red}{a modular pixel-level feature representation module is developed} that employs self-knowledge distillation. The Pixel-level Generation Module (PGM) directly integrates object cues from diverse modalities, effectively mitigating noise interference, as visualized in Fig. 3. In the distillation learning process, a teacher network using the SeAFusion method\cite{55} is engaged, with the PGM functioning as the student network, extracting knowledge from the teacher. \textcolor{red}{The SeAFusion model mainly includes a fusion module and a semantic segmentation module, which have a strong ability for feature representation. There are, of course, our teacher network can also be replaced with other high-performance models. The underlying concept of the student network is centered on lightweight efficiency; accordingly, it preserves the architectural design of the teacher network while substantially diminishing the parameterization within the hidden layers of the neural network.}

\begin{algorithm}
\caption{The training process of PGM.}\label{algo1}
\begin{algorithmic}[1]
\Require RGB and T images
\Ensure fused image F
\State $epoch \leftarrow 1$
\While{$epoch \leq max\_epoch$}
    \State $iteration \leftarrow 1$
    \While{$iteration < the\_number\_of\_iteration$}
        \State Select $m$ RGB images $\{ I_{rgb}^1, \cdots ,I_{rgb}^m\}$
        \State Select $m$ T images $\{ I_t^1, \cdots ,I_t^m\}$
        \State Update output images of the teacher network $\{ O_{te}^1, \cdots ,O_{te}^m\}$
        \State Select $m$ RGB images $\{ I_{rgb}^1, \cdots ,I_{rgb}^m\}$
        \State Select $m$ T images $\{ I_t^1, \cdots ,I_t^m\}$
        \State Add noise interference $N$
        \State Update the PGM by Adam Optimizer
        \State $iteration \leftarrow iteration + 1$
    \EndWhile
    \State $epoch \leftarrow epoch + 1$
\EndWhile
\State \Return fused image F
\end{algorithmic}
\end{algorithm}

\begin{figure*}
    \centering
    \includegraphics[width=1.0\linewidth]{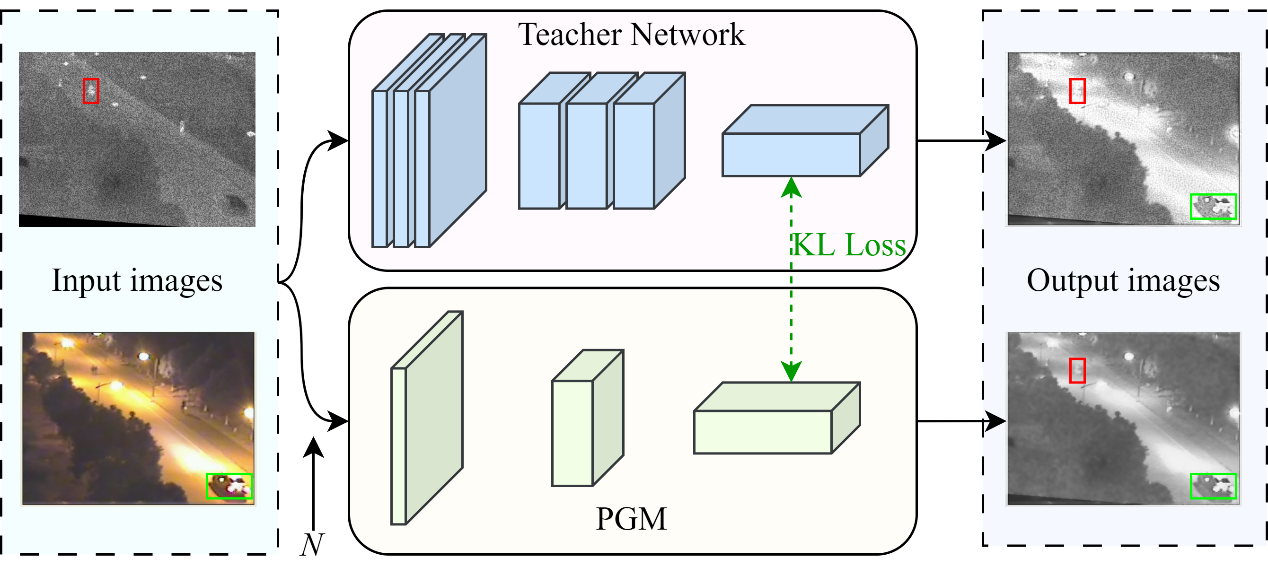}
    \caption{The details of PGM. $N$ denotes the noise interference.}
    \label{fig3}
\end{figure*}
\begin{figure*}
    \centering
    \includegraphics[width=0.85\linewidth]{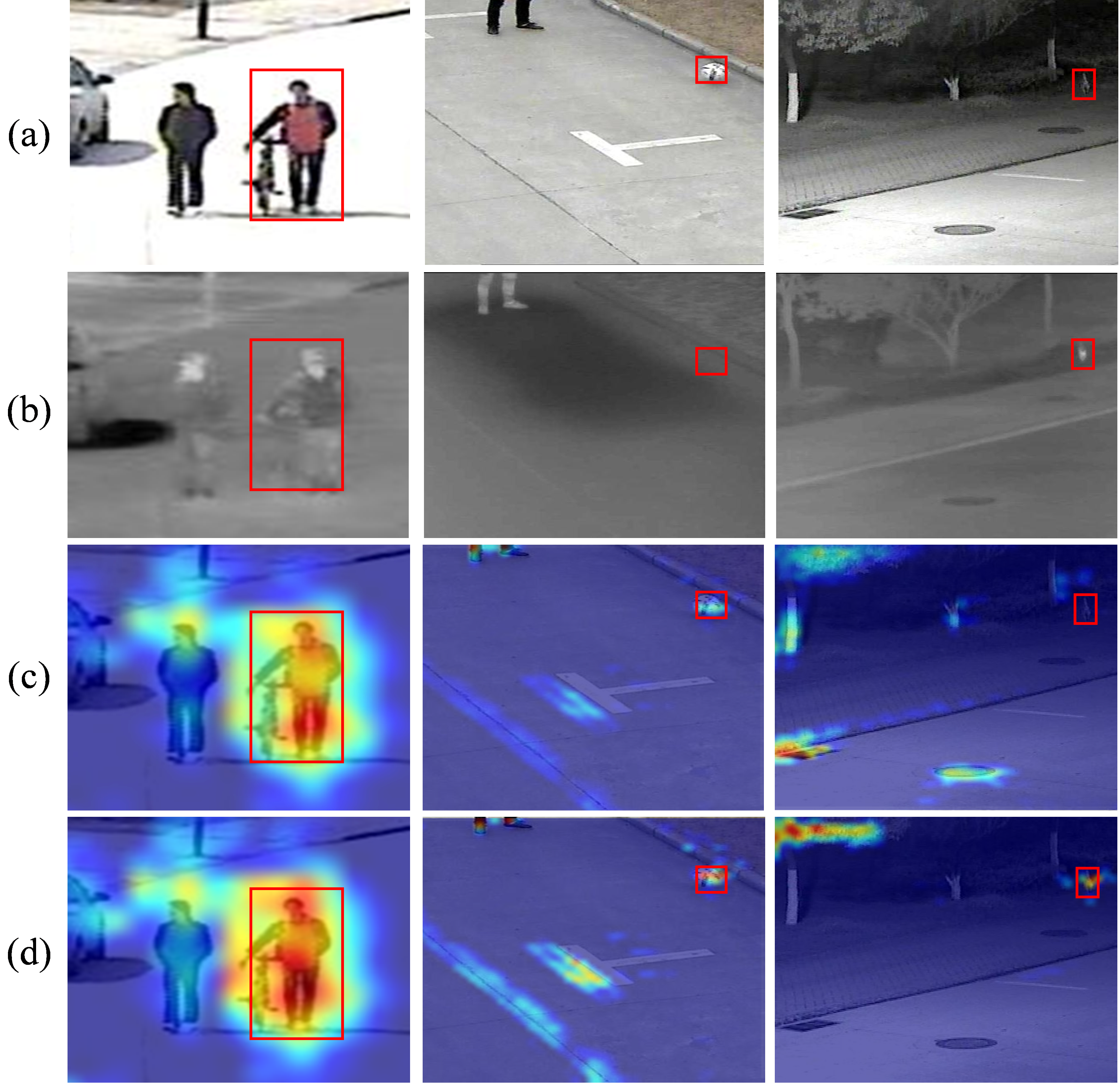}
    \caption{Illustration of the effectiveness of feature attention in FIM. (a) RGB images, (b) Thermal images, heatmap of features (c) without FIM, (d) with FIM.}
    \label{fig4}
\end{figure*}
The training procedure for the PGM is detailed in Algorithm 1. The architecture of PGM is built on a minimalist set of convolutional and deconvolutional layers. Despite its compact nature, which boasts fewer parameters and lower computational costs than the teacher network, the PGM delivers performance on par. \textcolor{red}{Inspired by the theory of feature enhancement and the requirement for interference resistance in neural networks, a Gaussian noise-based mask has been designed for the student network during the knowledge distillation process. This mask is intended to compel the PGM to learn more effective depth features, thereby achieving fusion results with a high signal-to-noise ratio (SNR). The introduction of the Gaussian noise mask serves as a regularization technique that encourages the PGM to focus on the most discriminative aspects of the data while ignoring irrelevant or noisy information, ultimately leading to a more robust and generalized model capable of extracting salient features under varying conditions.} Consequently, the outputs of PGM show a superior ability to reduce noise when compared to those of the teacher network. The advantages of self-knowledge distillation enable the PGM to meld modality-specific features effectively, while maintaining rich texture details and highlighting thermal object characteristics.

\textbf{Feature-level interaction module (FIM).} The resultant fused image X constitutes a pivotal element in the synthesis of multi-modal representation, amalgamating the complementary information contained within RGB-T images. This process is effective in noise interference mitigation. Following feature extraction, the RGB, T, and X source images are inputted into a feature extraction module, yielding deep features ${D{rgb}},{Dt},{D_x} \in {\mathbb{R}^{H \times W \times C}}$, wherein $H$, $W$, and $C$ denote the image height, width, and channel number, respectively. Subsequently, \textcolor{red}{to enhance feature representation capabilities without increasing the number of learning parameters, the FIM is designed consisting of two basic modules.} To fully capture the channel-wise correlations among deep features, the Spatial Feature Transformation and Shifting (SFTS) module is proposed to segment the deep features ${D{rgb}}$ and ${Dt}$ into four slices across the channel plane. These slices are then shifted by a single pixel in each of the four cardinal directions, as described below:

\begin{equation}
\label{deqn_ex1a1}
\left\{ \begin{gathered}
  \overline {{D_{rgb}}} [:,2:W,1:C/4] \leftarrow {D_{rgb}}[:,1:W - 1,1:C/4] \hfill \\
  \overline {{D_{rgb}}} [:,1:W-1,C/4{\text{ + }}1:C/2] \leftarrow {D_{rgb}}[:,2:W,C/{\text{4 + }}1:C/2] \hfill \\
  \overline {{D_{rgb}}} [2:H,:,C/2:3C/4] \leftarrow {D_{rgb}}[1:H - 1,:,C/2:3C/4] \hfill \\
  \overline {{D_{rgb}}} [1:H - 1,:,3C/4:C] \leftarrow {D_{rgb}}[2:H,:,3C/4:C] \hfill \\ 
\end{gathered}  \right.
\end{equation}

\begin{equation}
\label{deqn_ex1a2}
    \left\{ \begin{gathered}
  \overline {{D_t}} [2:H,:,1:C/4] \leftarrow {D_t}[1:H - 1,:,1:C/4] \hfill \\
  \overline {{D_t}} [1:H - 1,:,C/4{\text{ + }}1:C/2] \leftarrow {D_t}[2:H,:,C/{\text{4 + }}1:C/2] \hfill \\
  \overline {{D_t}} [:,2:W,C/2:3C/4] \leftarrow {D_t}[:,1:W - 1,C/2:3C/4] \hfill \\
  \overline {{D_t}} [:,1:W - 1,3C/4:C] \leftarrow {D_t}[:,2:W,3C/4:C] \hfill \\ 
\end{gathered}  \right.
\end{equation}

The SFTS introduces a spatial shift to the elements in order to avoid excessive interference due to object movement. Moreover, to prevent information disorder and visual distortion caused by excessive displacement, the SFTS module is designed to shift elements by a single pixel spatially. However, when two objects are in close spatial proximity, their features might intersect or overlap, potentially causing tracking failures. In response, the Mixed Feature Interaction Transformer (MFIT) is designed, which integrates the features produced by the SFTS to alleviate interference from spatial displacement.

The MFIT is based on self-attention and cross-modal attention. Initially, the features $\overline {{D_{rgb}}}$ and $\overline {{D_t}}$ are reshaped to obtain their own \textit{Query}  $\{ {q^{rgb}},{\text{ }}{q^t} \in {\mathbb{R}^{HW \times C}}\} $, \textit{Key} $\{ {k^{rgb}},{\text{ }}{k^t} \in {\mathbb{R}^{HW \times C}}\} $ and \textit{Value} $\{ {v^{rgb}},{\text{ }}{v^t} \in {\mathbb{R}^{HW \times C}}\}$, respectively. Subsequently, the cross-modal attention map is calculated by the following formulas:

\begin{equation}
    {D_{rgb - t}} = \operatorname{softmax} \left( {\frac{{{q^t}{{({k^{rgb}})}^{\text{T}}}}}{{\sqrt {{d_k}} }}} \right){v^{rgb}}
\end{equation}

\begin{equation}
    {D_{t - rgb}} = \operatorname{softmax} \left( {\frac{{{q^{rgb}}{{({k^t})}^{\text{T}}}}}{{\sqrt {{d_k}} }}} \right){v^t}
\end{equation}
where $\{ {D_{rgb - t}},{\text{ }}{D_{t - rgb}} \in {\mathbb{R}^{H \times W \times C}}\}$ stands for a pair of attention maps, and ${d_k}$ denotes the scaling factor, set as 1. Ultimately, adding the attention maps and linking with the deep fusion features ${D_X}$ to obtain the informative fusion maps $\overline {{D_f}}  \in {\mathbb{R}^{H \times W \times 2C}}$.
\begin{equation}
    At{t^{rgbt}} = {D_{rgb - t}} + {D_{t - rgb}}
\end{equation}
\begin{equation}
    \overline {{D_f}}  = {\text{concat}}({D_x},At{t^{rgbt}})
\end{equation}

The feature visualization of the FIM is shown in Fig. 4, illustrating that the integration of FIM enables the tracker to focus effectively on small objects, thereby enhancing its ability to concentrate on minute target regions. Conversely, the absence of FIM might result in a failure to adequately focus on these small target regions.

Following fusion, the feature maps are cropped using RoIAlign and provided to a binary classifier for the determination of a coarse location. In scenarios involving abrupt drifts between successive frames, the local search approach of the base tracker is shown to be inadequate. Furthermore, the initial frame-based learning of bounding box regression by the base tracker is unable to dynamically adapt the target scale in subsequent frames, rendering dynamic scale adjustment a challenging task.

\textbf{Decision-level refinement module (DRM).} \textcolor{red}{A flexible decision-level refinement strategy is proposed by combining optical flow and a refinement network, which estimates motion offsets and refines coarse positioning. The decision-level refinement module optimizes tracking performance by improving the precision of the predictions. Furthermore, since this module determines whether to use the optical flow method or the refinement network for re-localization based on the confidence score, it ensures both tracking efficiency and accuracy, in contrast to methods that rely solely on either the optical flow method or the refinement network.} 

The DRM implements different bounding box regression strategies based on the confidence score ${C_S}$. When ${C_S} < 0$, DRM applies the Lucas-Kanade optical flow\cite{29} repositioning rule based on pyramid layering to obtain the target motion offset. The final target positioning is achieved by adding the target location obtained from the local search using Gaussian sampling to the offset in motion. If ${C_S} > 0$, the refinement network is initialized to consistently improve the accuracy of the predicted bounding box. In particular, we fine-tune the plug-and-play Alpha Refine\cite{30} component on the RGBT dataset, which can refine the bounding boxes accurately with its exceptional spatial awareness capability.

\subsection{Network Training}
During the offline training phase, we utilize the VGG-M architecture as our backbone and implement a multi-domain learning strategy. Afterward, the proposed method randomly samples 8 frames and extracts 32 positive samples and 96 negative samples per frame using Intersection over Union (IoU) and Gaussian distribution. These samples are employed to construct a mini-batch. \textcolor{red}{The XNet is optimized by adopting} the AdamW and setting the learning rate for the convolution layers to $1{e^{ - 4}}$, for the fully connected layers to $1{e^{ - 4}}$, and the training epoch to 200.

\subsection{Online Tracking}

\begin{table}
    \centering
    \normalsize
\begin{tabularx}{1.0\textwidth}{l}
\hline
\textbf{Algorithm 2.} The online tracking process of X-Net.             \\
\hline
\textbf{Input:} All pre-trained parameters of the backbone.                              \\
Original target position ${Z_1}$.\\
\textbf{Output:} Predicted target position $Z_t^*$                                \\
1: Initialized the parameters of ${{\mathop{\rm W}\nolimits} _{fc6}}$;\\
2: Train the bounding box regressor $BR( \cdot )$;\\
3: Generate positive and negative samples $S_1^ + ,S_1^ - $;\\
4: Renew ${{\rm{W}}_{{\rm{fc4 - fc6}}}}$ through $S_1^ + ,S_1^ - $;\\
5: Generate short-term and long-term samples ${U_s},{U_l}$.\\
6: \textbf{repeat}\\
7: \quad Draw candidate samples $z_t^i$.\\
8: \quad Calculate the optimum target state $Z_t^*$.\\
9: \quad \textbf{if} ${f^ + }\left( {Z_t^*} \right) > 0.5$ \textbf{then}\\
10: \quad\quad   Generate training samples $S_t^ + ,S_t^ - $;\\
11: \quad\quad   Computing  $Z_t^* = BR(Z_t^*)$;\\
12: \quad\quad   Update ${U_s},{U_l}$.\\
13: \quad \textbf{if} ${f^ + }\left( {Z_t^*} \right) < 0.5$ \textbf{then}\\
14: \quad\quad   Execute short-term update ${{\mathop{\rm W}\nolimits} _{fc4 - fc6}}$ with ${U_s}$.\\
15: \quad \textbf{else if} $t\% 10 = 0$ \textbf{then}\\
16:\quad\quad    Execute long-term update ${{\mathop{\rm W}\nolimits} _{fc4 - fc6}}$ with ${U_l}$.\\
17: \quad \textbf{if} ${C_S} > 0$ \textbf{then}\\
18: \quad\quad   Obtain the refine target position $Z_m^*$.\\
19: \quad \textbf{else if} ${C_S} < 0$ \textbf{then}\\
20: \quad\quad   Calculate the average displacement vector $[{x^*},{y^*}]$.\\
21: \quad\quad   Obtain the predict target position $Z_m^*$.\\
22:\quad  Improve $Z_m^*$ to $Z_t^*$.\\
23: \textbf{until} termination of sequence.\\
\hline
\end{tabularx}
\end{table}

\textbf{Algorithm 2} delineates the comprehensive online tracking mechanism. Specifically, the fully connected (FC) layers within the backbone are updated, while all other parameters are kept constant. From the starting gate, X-Net is initialized on the first frame of the sequence with the target position. Gaussian sampling is performed around the target in the first frame to obtain 500 positive samples $S_1^ + $ with $\operatorname{IoU}  > 0.7$, and 5000 negative samples $S_1^ - $ with $\operatorname{IoU}  < 0.5$ w.r.t the ground truth. The FC layers undergo 50 epochs of fine-tuning based on these samples, with a learning rate of $1{e^{ - 4}}$ for FC6 and $1{e^{ - 3}}$ for the remaining layers. A total of 1000 samples are chosen for training the bounding box regressor with the objective of obtaining a precise bounding box that encompasses the target. 

In order to mitigate the impact of tracking inefficiency and inaccurate predictions for subsequent video frame tracking boxes, the regressor is trained only using the initial frame and the corresponding bounding box. The regression model modifies the tracking coordinates of subsequent frames to achieve more accurate tracking. Thereafter, both short- and long-term updates are conducted using a set of 50 positive samples $S_t^ + $ and 200 negative samples $S_t^ - $. A Gaussian sampling technique is utilized with $Z_{t - 1}^n$ as the center to generate 256 candidate regions $Z_t^i$ for the $t$th frame. The trained network evaluates the positive scores ${f^ + }\left( {Z_t^i} \right)$ and negative scores ${f^ - }\left( {Z_t^i} \right)$ for each of these candidate samples. Subsequently, the confidence scores of the candidates are sorted and the top 5 candidates with the highest scores are selected and averaged as the set $Z_m^*$. Ultimately, the DRM is employed as the optimization metric for refining the bounding box.

\section{Experimental Results}
The X-Net is implemented on the PyTorch 1.10 platform with one NVIDIA RTX3090 GPU with 24GB memory.

\subsection{Datasets and Metrics}
In this paper, \textcolor{red}{the comparative experiments are conducted with high-performance competitors on three RGBT benchmarks}, namely GTOT\cite{31}, RGBT234\cite{32}, and LasHeR\cite{33}. In alignment with prevalent practices within the domain, two benchmark metrics, Precision Rate (PR) and Success Rate (SR), are adopted to ascertain the efficacy of the proposed method. A threshold of 5 pixels is established for GTOT and 20 pixels for RGBT234/LasHeR, taking into account the variability in image resolutions across different datasets.

\subsection{Evaluation on GTOT Dataset}
1) \textbf{Overall performance.} The proposed tracker is compared with 11 state-of-the-art RGBT methods, i.e., MIRNet, APFNet, AGMINet, MFGNet\cite{34}, SiamCDA, HDINet\cite{35}, MSIFNet\cite{36}, RPCF\cite{37}, M2GCI\cite{38}, EDFNet\cite{39} and DMSTM\cite{40}. \textcolor{red}{A succinct overview of the comparisons is presented in Table \ref{table1}.} As reported in Table \ref{table2}, X-Net exhibits superior tracking capability compared to the state-of-the-art trackers, with PR of 93.1$\%$, surpassing the compared trackers by approximately 0.4$\%$-12.8$\%$. Moreover, with regard to the SR, X-Net outperforms compared trackers, obtaining an impressive SR of 76.7$\%$. The results demonstrate that X-Net is capable of attaining exceptional performance.

\begin{table}[htbp]
\caption{\textcolor{red}{The introduction of compared trackers.}}
\begin{tabular}{llll}
\hline
Name & Pub. Info                                                                               & Year & Categorize             \\ \hline
MIRNet   & ICME           & 2022 & Generative Tracker     \\
APFNet   & AAAI     & 2022 & Generative Tracker     \\
AGMINet  & IEEE TIM                                  & 2022 & Generative Tracker     \\
HDINet   & IEEE Sensor journal                                                                     & 2021 & Generative Tracker     \\
MFGNet   & IEEE TMM                                                        & 2022 & Generative Tracker     \\
SiamCDA  & IEEE   TCSVT & 2023 & Discriminative Tracker \\
RPCF     & Expert Systems with Applications                                                        & 2023 & Discriminative Tracker \\
MSIFNet  & Sensors                                                                                 & 2023 & Generative Tracker     \\
HATFNet  & Applied intelligence                                                                    & 2023 & Generative Tracker     \\
DMSTM    & IEEE TIM                                  & 2023 & Generative Tracker     \\
EDFNet   & The Journal of Supercomputing                                                           & 2023 & Generative Tracker     \\
M2GCI    & Neural Processing Letters                                                               & 2023 & Generative Tracker    \\
\hline
\end{tabular}
    \label{table1}
\end{table}

\begin{table*}[htbp]
\caption{Comparison results of the proposed method against the state-of-the-art trackers. Overall performance is evaluated by PR/SR scores ($\%$) and is produced on GTOT. The best results are marked in \textcolor{red}{red}.}
\label{tab1}
\scalebox{0.93}{\begin{tabular}{ccccccc}
\hline
Trackers & MIRNet    & APFNet    & AGMINet   & MFGNet    & SiamCDA   & HDINet    \\
ALL      & 90.9/74.4 & 90.4/73.5 & 91.7/73.4 & 80.3/65.3 & 87.7/73.2 & 88.8/71.8  \\ \hline
Trackers& MSIFNet   & RPCF      & M2GCI     & EDFNet    & DMSTM     & X-Net  \\ 
ALL& 90.4/74.1 & 92.7/64.6 & 90.9/73.7 & 91.2/74.6 & 92.9/75.9 & \textcolor{red}{93.1}/\textcolor{red}{76.7} \\ \hline
\end{tabular}}
\label{table2}
\end{table*}

2)	 \textbf{Challenge-based performance.} To evaluate the superiority and robustness of the designed methodology, the challenge-based performance tests are conducted on the GTOT dataset and compared with 11 excellent RGBT trackers, including CAT\cite{41}, APFNet, ADRNet, HMFT\cite{42}, JMMAC\cite{43}, MaCNet, MIRNet, BACF\cite{44}, ECO\cite{45}, RT-MDNet\cite{46} and SiamDW\cite{47}. The visualization of the results of the attribute challenge is shown in Fig. 5, and the lines represent the Euclidean distance between the predicted centroid point of the compared methods and the ground truth. It can be observed that the prediction of X-Net is always closer to the ground truth bounding box than the comparisons, regardless of whether facing scale variation (SV) and fast motion (FM) challenges. 

\begin{figure*}
    \centering
    \includegraphics[width=1.0\linewidth]{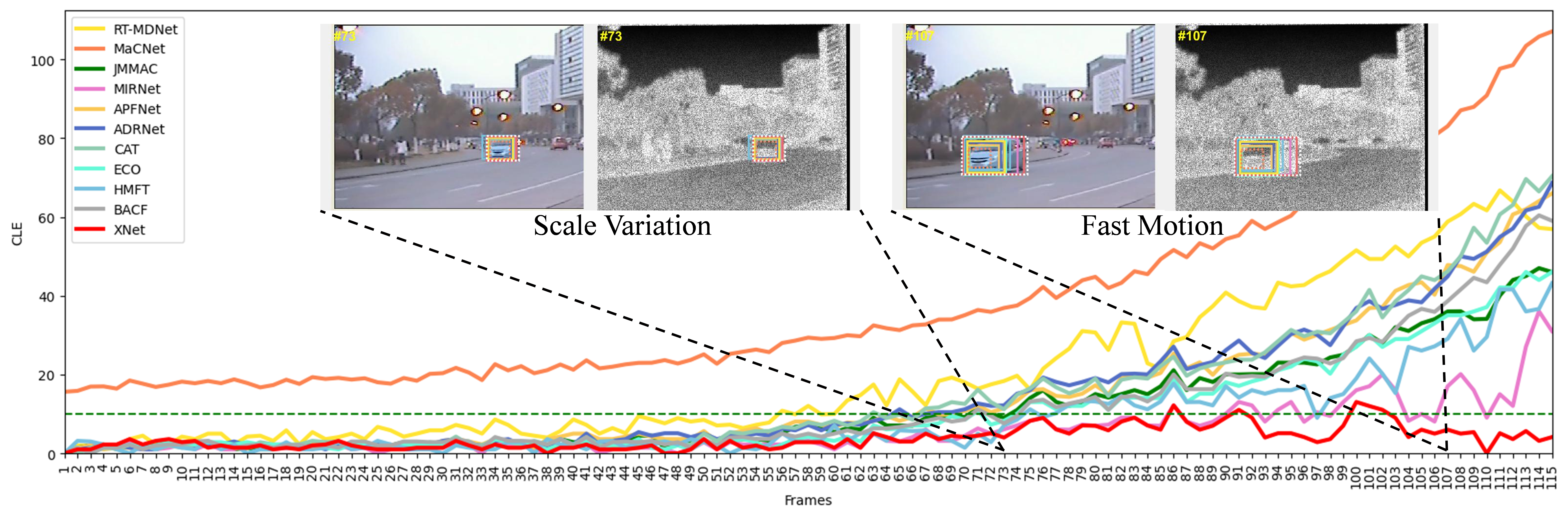}
    \caption{Comparison results of X-Net against the state-of-the-art trackers. Challenge-based performance is evaluated by PR/SR scores ($\%$) and produced on GTOT.}
    \label{fig5}
\end{figure*}
\begin{figure*}[htbp]
    \centering
    \includegraphics[width=1.0\linewidth]{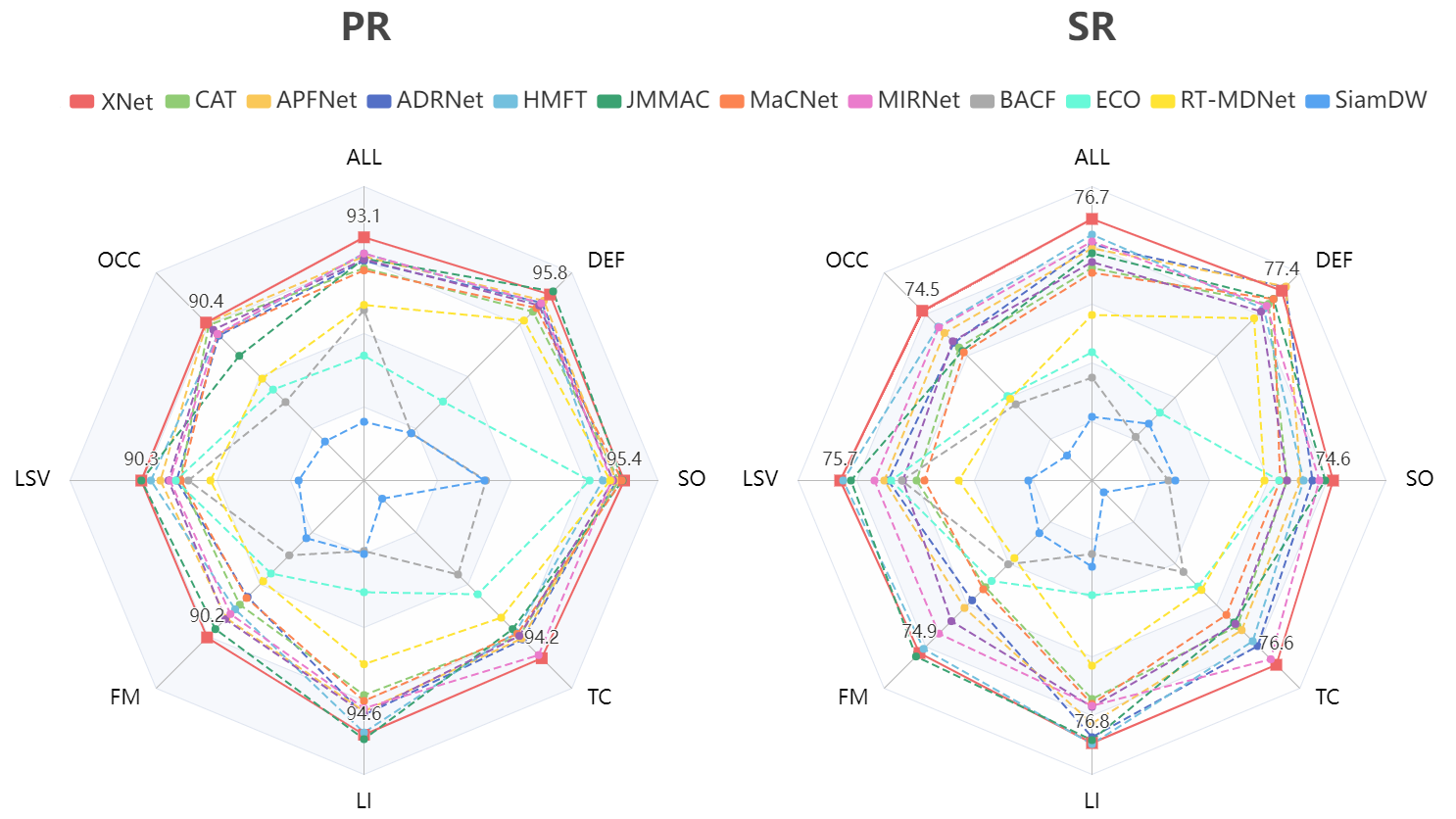}
    \caption{Visualization of the results of attribute challenge.}
    \label{fig6}
\end{figure*}

In Fig. 6, the tracking performance is evaluated across seven challenges: deformation (DEF), large scale variation (LSV), occlusion (OCC), fast motion, low illumination (LI), thermal crossover (TC), and small object (SO). The proposed method clearly excels in OCC, LSV, LI, TC, and SO challenges, as indicated by both PR and SR metrics, which suggests that X-Net possesses exceptional resistance to interference and versatility across various scenarios. Furthermore, X-Net demonstrates optimal or near-optimal performance in addressing DEF and FM challenges, illustrating its capability for precise target tracking despite the target undergoing diverse transformations. In the radar chart visualization of the GTOT dataset, the proposed tracker X-Net presents a distinct octagonal shape, providing a more comprehensive performance comparison relative to other methods. This figure underscores the capability of X-Net to handle multiple challenges with outstanding performance and highlights its remarkable robustness.

\begin{figure*}
    \centering
    \includegraphics[width=1.0\linewidth]{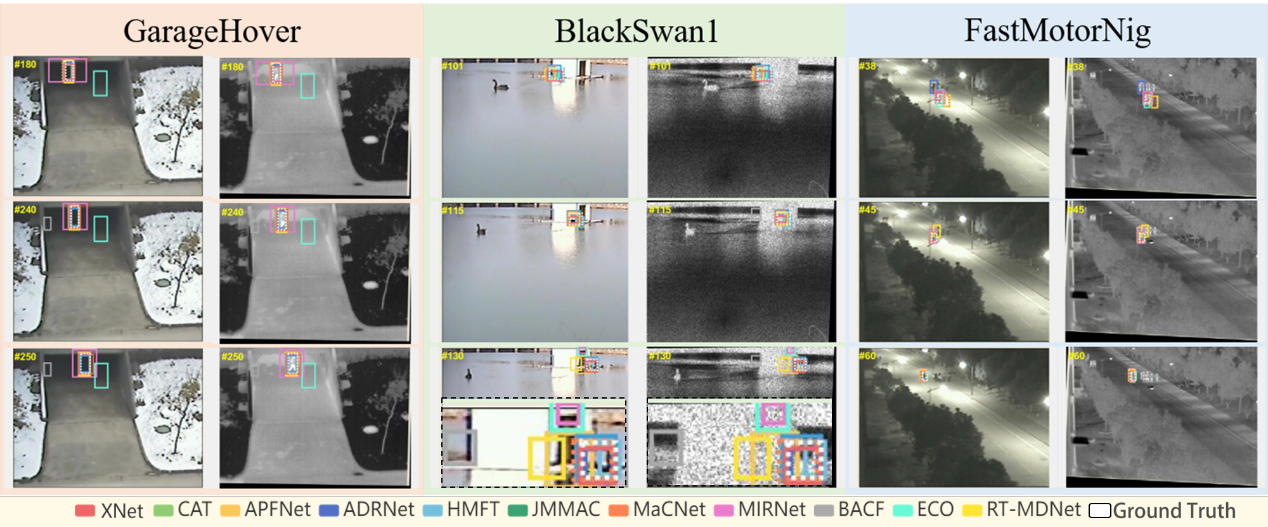}
    \caption{Qualitative comparison results of different trackers on the GTOT dataset.}
    \label{fig7}
\end{figure*}

3)	\textbf{Qualitative comparison.} In order to comprehensively showcase the benefits of X-Net, a qualitative comparison of the tracking results on \textit{GarageHover}, \textit{BlackSwan1} and \textit{FastmotorNig} sequences is conducted with 10 exemplary trackers, as shown in Fig. 7. Specifically, the \textit{GarageHover} sequence reveals the inability of the ECO and BACF methods to track the object in low illumination conditions. The tracking boxes of MIRNet are noticeably inaccurate. In the case of the \textit{BlackSwan1} sequence, the zoomed-in section of the tracking results for the 130th frame is presented in the bottom black box of the third row in the image, which illustrates the BACF, ECO, RT-MDNet, APFNet and MIRNet methods struggle to track the object when faced with background interference. In scenarios where tracking occurs during nighttime with rapid target movement (\textit{FastMotorNig}), conventional methods commonly encounter challenges, such as imprecise tracking or even misidentifying targets. In contrast, the proposed RGBT tracker consistently achieves stable and accurate tracking performance.

\subsection{Evaluation on RGBT234 Dataset}

1)	\textbf{Overall performance.} The RGBT234 dataset is applied to evaluate the performance comprehensively, as reported in Table \ref{table3} and Table \ref{table4}. It is evident that the X-Net outperforms the comparisons both in the PR and SR metrics. The proposed method surpasses the EDFNet method by 2.5$\%$ and 4.2$\%$ in terms of PR and SR, respectively. Significantly, X-Net demonstrates an improvement of approximately 17.1$\%$/14$\%$ in PR/SR, compared to RPCF.
\begin{table*}[!htbp]
\caption{Comparison results of the proposed method against the state-of-the-art trackers. attribute-based and overall performance are evaluated by PR scores ($\%$) and are produced on RGBT234. The best results are marked in \textcolor{red}{red}.}
\scalebox{0.8}{
\begin{tabular}{cccccccccccccc}
\hline
Trackers & NO   & PO   & HO   & LI   & LR   & TC   & DEF  & FM   & SV   & MB   & CM   & BC   & ALL  \\ \hline
MIRNet   & 95.4 & 86.1 & 71.0 & 83.4 & 83.9 & 81.1 & 77.8 & 68.3 & 82.7 & 74.6 & 76.4 & 78.9 & 81.6 \\
APFNet   & 94.8 & 86.3 & 73.8 & 84.3 & 84.4 & 82.2 & 78.5 & 79.1 & 83.1 & 74.5 & 77.9 & 81.3 & 82.7 \\
AGMINet  & 94.9 & 90.2 & 72.9 & \textcolor{red}{87.0} & \textcolor{red}{86.7} & 80.6 & 79.5 & \textcolor{red}{79.4} & 83.2 & 78.2 & 79.0 & \textcolor{red}{83.3} & 84.0 \\
MFGNet   & 92.0 & 84.3 & 66.2 & 79.1 & 79.3 & 81.8 & 72.1 & 72.5 & 76.1 & 73.7 & 73.2 & 74.3 & 78.3 \\
SiamCDA  & 88.4 & 84.2 & 66.2 & 81.8 & 70.9 & 67.4 & 77.9 & 61.4 & 77.7 & 63.6 & 73.3 & 74.0 & 76.0 \\
HDINet   & 88.4 & 84.9 & 67.1 & 77.7 & 80.1 & 77.2 & 76.2 & 71.7 & 77.5 & 70.8 & 69.7 & 71.1 & 78.3 \\
MSIFNet  & 92.6 & 84.7 & 73.8 & 83.2 & 85.1 & 84.4 & 74.7 & 74.7 & 82.0 & 75.8 & 77.3 & 80.5 & 81.7 \\
RPCF     & 81.5 & 71.8 & 58.3 & 64.4 & 62.5 & 64.8 & 65.5 & 55.2 & 70.6 & 58.0 & 61.1 & 56.6 & 68.1 \\
HATFNet  & 92.1 & 86.0 & 70.3 & 78.4 & 79.6 & 75.5 & 77.4 & 74.0 & 80.9 & 71.1 & 76.7 & 77.9 & 80.7 \\
M2GCI    & 88.0 & 82.8 & 71.9 & 75.2 & 78.3 & 74.0 & 75.1 & 69.1 & 80.2 & 70.0 & 69.0 & 77.1 & 79.3 \\
EDFNet   & 91.5 & 85.3 & \textcolor{red}{76.2} & 78.4 & 80.4 & 75.2 & 79.9 & 72.1 & 83.8 & 73.6 & 73.6 & 80.3 & 82.7 \\
DMSTM    & 90.5 & 79.0 & 72.9 & 78.8 & 75.2 & 72.2 & 80.0 & 76.0 & 54.0 & 72.9 & 76.5 & 72.4 & 78.6 \\
X-Net     & \textcolor{red}{94.9} & \textcolor{red}{90.5} & 75.7 & 83.2 & 84.8 & \textcolor{red}{86.9} & \textcolor{red}{82.4 }& 76.3 & \textcolor{red}{87.6} & \textcolor{red}{80.6} & \textcolor{red}{81.7} & 78.3 & \textcolor{red}{85.2} \\ \hline
\end{tabular}}
\label{table3}
\end{table*}
\begin{table*}[!htbp]
\caption{Comparison results of the proposed method against the state-of-the-art trackers. attribute-based and overall performance are evaluated by SR scores ($\%$) and are produced on RGBT234. The best results are marked in \textcolor{red}{red}.}
\scalebox{0.8}{
\begin{tabular}{cccccccccccccc}
\hline
Trackers & NO   & PO   & HO   & LI   & LR   & TC   & DEF  & FM   & SV   & MB   & CM   & BC   & ALL  \\ \hline
MIRNet   & 72.4 & 62.7 & 49.0 & 57.5 & 56.3 & 59.1 & 58.1 & 47.1 & 61.9 & 54.6 & 55.4 & 51.7 & 58.9 \\
APFNet   & 68.0 & 60.6 & 50.7 & 56.9 & 56.5 & 58.1 & 56.4 & 51.1 & 57.9 & 54.5 & 56.3 & 54.5 & 57.9 \\
AGMINet  & 69.1 & 63.9 & 50.3 & 59.8 & 57.2 & 59.2 & 56.8 & 51.2 & 59.3 & 57.5 & 57.5 & \textcolor{red}{55.3} & 59.2 \\
MFGNet   & 64.0 & 58.0 & 44.3 & 54.2 & 49.5 & 55.8 & 50.8 & 44.6 & 52.8 & 51.0 & 50.4 & 45.9 & 53.5 \\
SiamCDA  & 66.4 & 63.9 & 48.7 & \textcolor{red}{61.2} & 49.9 & 47.7 & 59.2 & 45.3 & 59.3 & 47.9 & 54.7 & 52.9 & 56.9 \\
HDINet   & 65.1 & 60.4 & 47.3 & 53.2 & 54.5 & 57.5 & 56.5 & 47.5 & 55.8 & 52.6 & 51.4 & 47.8 & 55.9 \\
MSIFNet  & 66.0 & 59.5 & 50.5 & 55.7 & 56.5 & 59.0 & 53.5 & 47.5 & 57.6 & 54.4 & 54.9 & 52.3 & 57.0 \\
RPCF     & 59.7 & 51.0 & 40.4 & 45.2 & 41.3 & 47.1 & 46.5 & 37.7 & 51.2 & 41.7 & 44.0 & 36.9 & 48.2 \\
HATFNet  & 70.3 & 64.2 & 50.0 & 55.2 & 55.1 & 56.2 & 58.4 & 51.4 & 61.2 & 53.9 & 56.9 & 52.4 & 59.5 \\
M2GCI    & 64.9 & 59.8 & 50.4 & 51.7 & 53.8 & 55.3 & 54.8 & 44.8 & 58.4 & 51.8 & 50.9 & 51.5 & 56.9 \\
EDFNet   & 66.1 & 60.9 & 51.4 & 52.8 & 54.9 & 56.3 & 55.9 & 45.7 & 59.6 & 52.8 & 51.9 & 52.6 & 58.0 \\
DMSTM    & 66.5 & 57.2 & 50.4 & 55.7 & 51.0 & 50.7 & 58.9 & 51.6 & 61.1 & 52.7 & 54.8 & 47.9 & 56.2 \\
X-Net     & \textcolor{red}{72.7} & \textcolor{red}{66.7} & \textcolor{red}{53.1} & 57.2 & \textcolor{red}{59.3} & \textcolor{red}{64.4} & \textcolor{red}{60.9} & \textcolor{red}{53.1} & \textcolor{red}{65.6} & \textcolor{red}{60.2} & \textcolor{red}{59.8} & 52.7 & \textcolor{red}{62.2} \\ \hline
\end{tabular}}
\label{table4}
\end{table*}

2)	\textbf{Attribute-based performance.} The challenging attributes of the RGBT234 dataset can be categorized into 12 types, i.e., no occlusion (NO), partial occlusion (PO), heavy occlusion (HO), low illumination (LI), low resolution (LR), thermal crossover (TC), DEF, fast motion (FM), scale variation (SV), motion blur (MB), camera movement (CM) and background clutter (BC). Based on the metric values presented in Table 2 and Table 3, X-Net achieves the highest levels of accuracy and success in the challenging attributes such as NO, PO, TC, DEF, SV, MB, CM and BC. X-Net surpasses the RT-MDNet-based method AGMINet and attains PR/SR gains of 0.3$\%$/2.8, 2.8$\%$/2.8$\%$, 6.3$\%$/5.2$\%$, 2.9$\%$/4.1$\%$, 4.4$\%$/6.3$\%$, 2.4$\%$/2.7$\%$, and 2.7$\%$/2.3$\%$ for the HO, TC, DEF, SV, MB, and CM challenges, respectively. 

Besides, X-Net exhibits outstanding performance, with impressive PR/SR scores of 90.5$\%$/66.7$\%$, 86.9$\%$/64.4$\%$, 80.6$\%$/60.2$\%$, and 81.7$\%$/59.8$\%$ when confronted with common attribute challenges including HO, TC, MB and CM, respectively. In the absence of no occlusions, the proposed tracker secures an elevated accuracy and success rate of 94.9$\%$/72.7$\%$. However, in the LI and BC challenges, X-Net exhibits inferior performance compared to AGMINet. This can be attributed to the ability of AGMINet to extract multi-scale information at each layer of features, allowing robust target localization even in the presence of low illumination and background clutter scenarios. The process of extracting features at each layer leads to an increase in computational complexity. X-Net exhibits exceptional competence in addressing attribute challenges, facilitating effective object tracking in diverse scenarios.

3)	\textbf{Qualitative comparison.} Qualitative comparison experiments are conducted on 9 video sequences, including \textit{aftertree}, \textit{baby}, \textit{carafetertree}, \textit{dog1}, \textit{manypeople}, \textit{nightthreepeople}, \textit{people1}, \textit{soccer2} and \textit{threeman2}, by comparing with 9 state-of-the-art trackers, as illustrated in Fig. 8. To facilitate the demonstration, we present the results using local magnification. 

Explicitly, in the \textit{aftertree} sequence, the target person is effectively tracked in all comparisons for the first 140 frames. However, at the 312th frame, when occluded by trees, the majority of trackers, specifically the ECO and HMFT methods, display tracking errors or inaccuracies. In the \textit{baby} sequence, the tracked target encounters attributes such as scale variation, occlusion, and thermal overlap as it approaches. The effects of tracking by different trackers are particularly notable. For example, most of the trackers have already encountered issues of tracking the wrong target at the 290th frame, and only MIRNet, ADRNet, and X-Net are able to track the target at the 537th frame. The target areas tracked by the MIRNet and ADRNet methods are significantly larger than the intended tracking area, while X-Net achieves the highest accuracy. Both camera movement and scale variation attributes are present in the sequences featuring \textit{caraftertree} and \textit{dog1}, leading to the difficulty of correctly tracking the target. The proposed method demonstrates accurate tracking of a car or dog, even during motion and partial occlusion. In the \textit{manypeople}, \textit{people1} and \textit{threeman2} sequences, the target faces challenges associated with occlusion and thermal overlap during the tracking process. 

\begin{figure*}[htbp]
    \centering
    \includegraphics[width=1.0\linewidth]{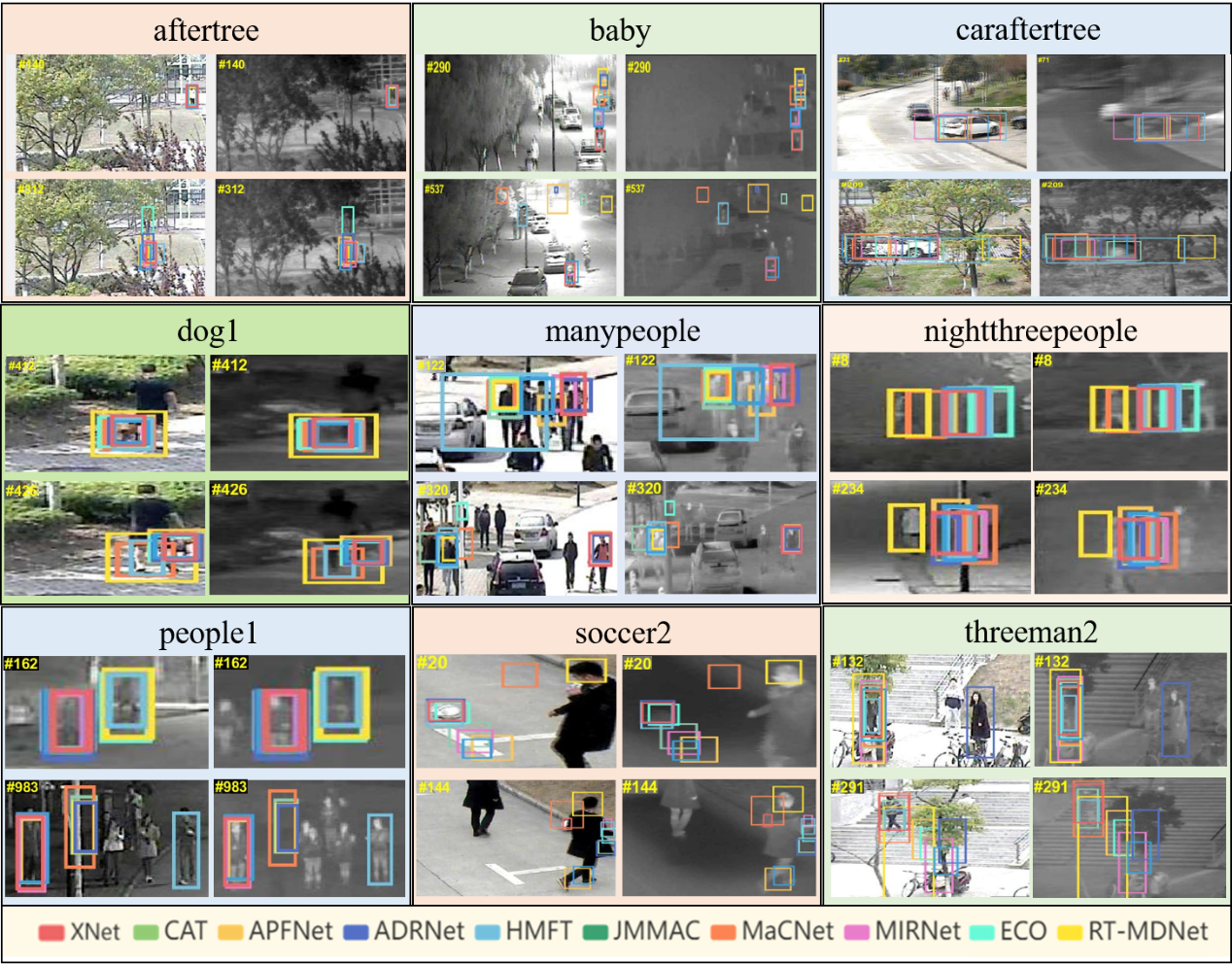}
    \caption{Qualitative comparison results of different trackers on the RGBT234 dataset.}
    \label{fig8}
\end{figure*}

Additionally, significant interference from surrounding persons adversely affects the performance of the trackers. The results from the two sequences demonstrate that X-Net consistently and accurately tracks the target without being affected by interference. When dealing with a rapidly moving soccer target, it becomes evident that most trackers struggle to accurately track the soccer within the first 20 frames. However, the proposed X-Net method proves to be the exception, successfully and precisely tracking the soccer by the 144th frame. While the other tracking methods being compared produce inaccurate results, X-Net consistently delivers precise and reliable tracking results.

In summary, the proposed method has consistently achieved exceptional tracking results on the challenging RGBT234 dataset, further confirming its superior tracking performance and robustness in various challenging scenarios. In medium-to-long video sequence tracking, X-Net surpasses the comparison methods, showcasing exceptional stability.

\subsection{Evaluation on LasHeR Dataset}
To further validate the robustness of X-Net, the model is trained on the RGBT234 dataset and evaluated on the testing subset of LasHeR. The comparison results of X-Net against 13 existing trackers, $i.e.$, APFNet, DMCNet, MaCNet, mfDiMP, MANet, CAT, DAPNet\cite{48}, MANet++, DAFNet\cite{49}, FANet\cite{50}, CMR\cite{51}, SGT++ and SGT\cite{52}, are shown in Fig. 9, which demonstrates the proposed method performs optimally in terms of PR and SR. Specifically, X-Net demonstrates superior performance over the SGT method with improvements of 18.1$\%$ and 20.4$\%$ in PR and SR metrics, respectively. Additionally, it achieves 4.1$\%$ and 10.2$\%$ higher PR and SR metrics compared to the MANet++ algorithm. In conclusion, the proposed X-Net demonstrates competitive tracking performance on LasHeR datasets, confirming its effectiveness, robustness, and ability to handle attribute challenges across diverse datasets.

\begin{figure}[htbp]
    \centering
    \includegraphics[width=1.0\linewidth]{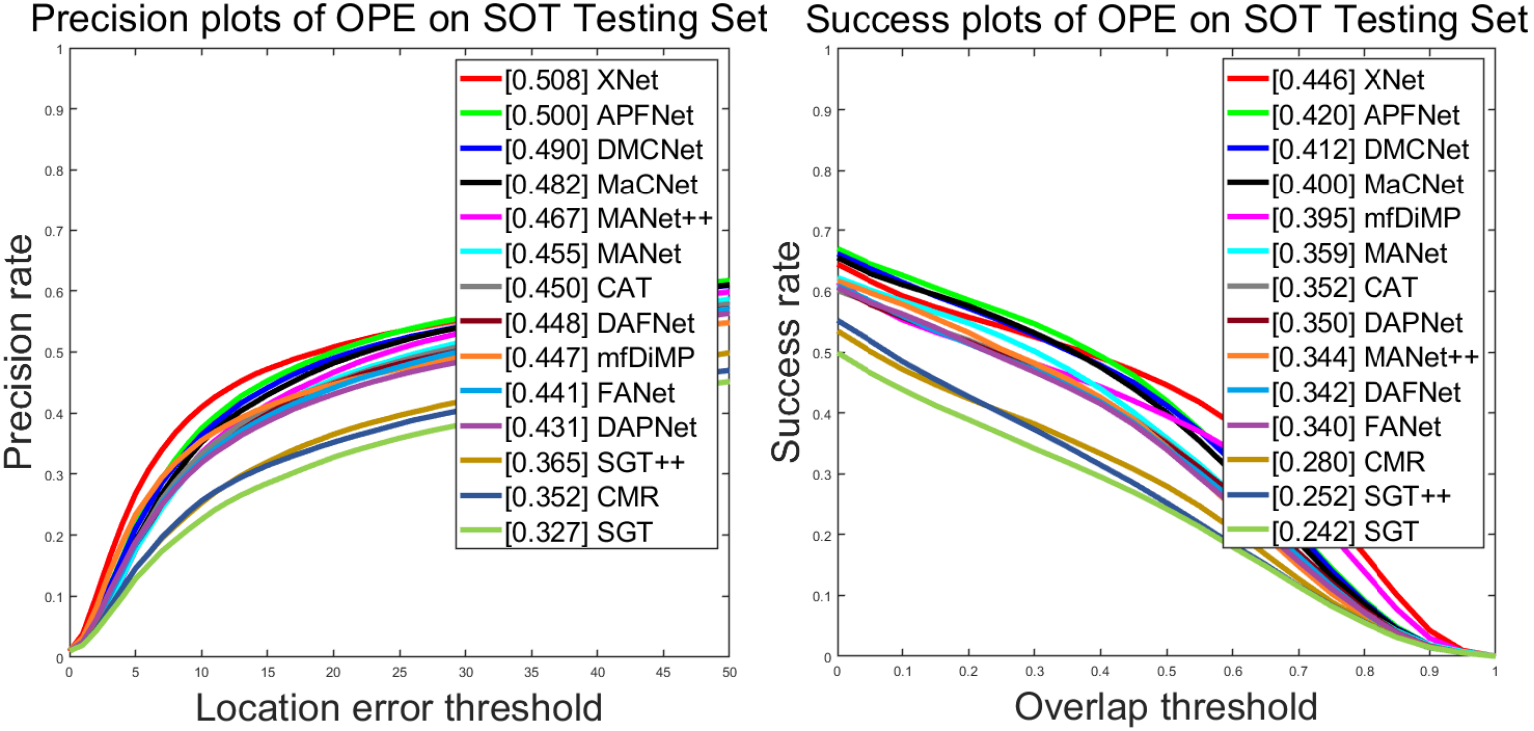}
    \caption{Comparison results of X-Net against the state-of-the-art trackers. Overall performance is evaluated by PR/SR scores ($\%$) and is produced on LasHeR.}
    \label{fig9}
\end{figure}

\subsection{Efficiency Analysis}
Tracking efficiency is critical for trackers and serves as a fundamental metric for their evaluation. In order to conduct a comprehensive evaluation of X-Net, the speed is benchmarked against six trackers on the GTOT dataset, as displayed in Fig. 10. It can be observed that X-Net demonstrates strong competitiveness in terms of performance and speed. Specifically, X-Net achieves a tracking speed of 21±3 fps, which are 7 and 9.5 times faster than those of the MANet and DMCNet, respectively. Notably, X-Net outperforms both DAFNet and MIRNet in terms of PR/SR metrics, even though it shows marginally slower execution speed.

\begin{figure}[htbp]
    \centering
    \includegraphics[width=1.0\linewidth]{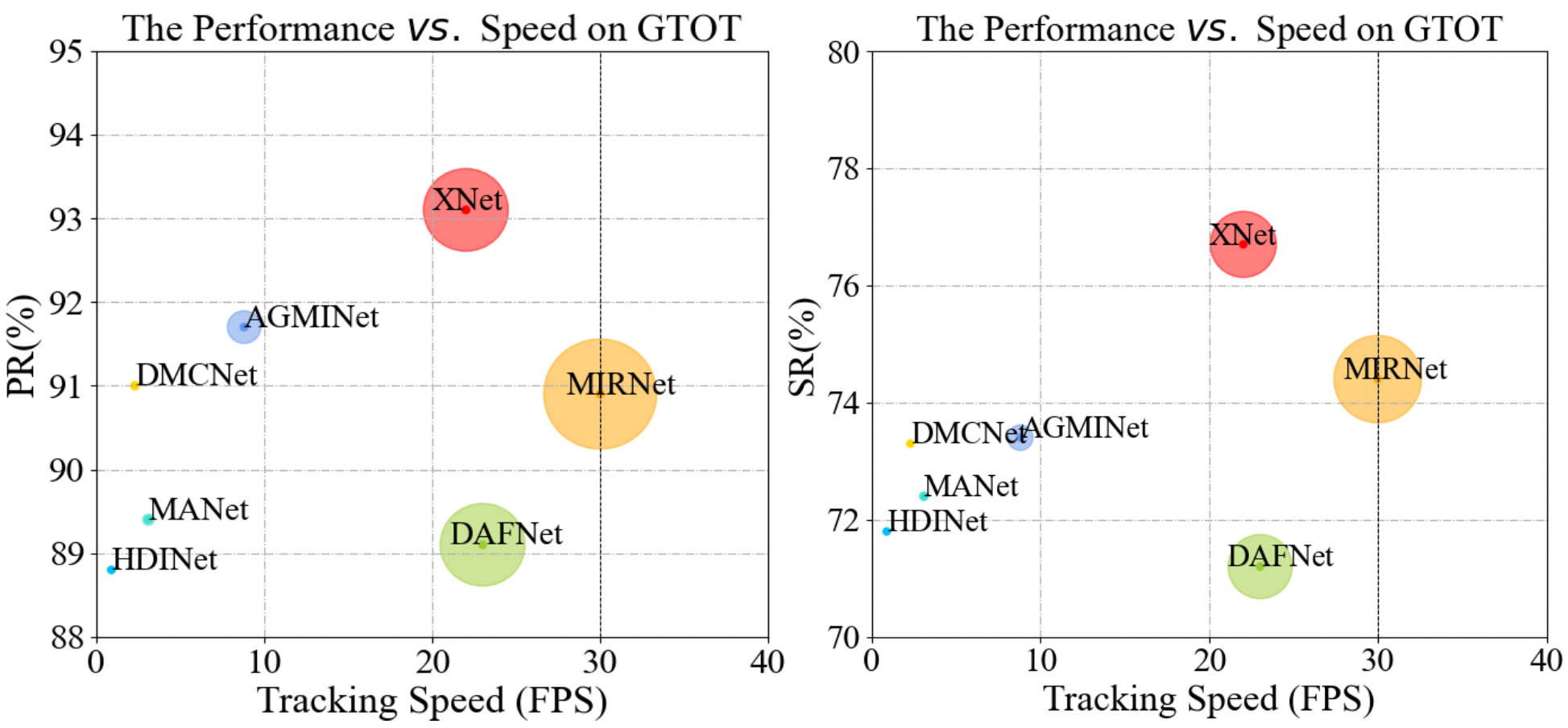}
    \caption{Illustration of the efficiency of X-Net against six trackers.}
    \label{fig10}
\end{figure}

\begin{table}[!ht]
\caption{Comparison results of the proposed method against the state-of-the-art trackers. attribute-based and overall performance are evaluated by PR scores ($\%$) and are produced on GTOT and RGBT234. The best results are marked in red.}
\begin{tabular}{cccccc}
\hline
 & MANet & HDINet & DMCNet & AGMINet      & X-Net   \\ \hline
Flops (GB)  & 0.28    & 0.27    & 0.94    & 0.35 & 0.68 \\
Params (MB)  & 7.11   & 9.40    & 11.45    & 10.76 & 6.93 \\
\hline
\end{tabular}
\label{table5}
\end{table}

Moreover, two crucial network metrics, namely Flops and Params, are compared to measure the complexity of the proposed model. Table \ref{table5} displays that the proposed method has lower parameters compared with the comparison methods, indicating the minimal space complexity of X-Net. It can be concluded that the X-Net model proposed in this study demonstrates lower Flops compared to DMCNet, showcasing its competitive advantage.

\subsection{Ablation Study}

\textcolor{red}{A sophisticated schema leveraging knowledge distillation techniques is implemented to transfer pre-eminently trained weights to the student network, designated as PGM, from a more complex progenitor network. An elucidative juxtaposition of the network parameters is delineated in Table \ref{table6}. Teacher-Net is the teacher network in the knowledge distillation architecture, referring to SeAFusion. Student-Net is the student network, which is PGM obtained through the process of knowledge distillation learning. The student network exhibits a reduction of 63.6$\%$ in FLOPs and 64.7$\%$  in parameters compared to the teacher network. Besides, the student network spends approximately 66.7$\%$ less time on the fusion of each pair of images compared to the teacher network. While simplifying the network architecture, the student network also achieved an upgrade in anti-interference capability by using Gaussian noise for data augmentation on the input data.}

\begin{table}[htbp]
\caption{\textcolor{red}{Comparison of computational complexity among PGM.}}
\begin{tabular}{cccc}
\hline
            & FLOPs(M) & Parameters(M) & times(s) \\\hline
Teacher-Net & 51.21     & 0.17          & 0.03     \\
Student-Net & 18.62    & 0.06          & 0.01   \\
\hline
\end{tabular}
\label{table6}
\end{table}

\begin{figure}[htbp]
    \centering
    \includegraphics[width=1.0\linewidth]{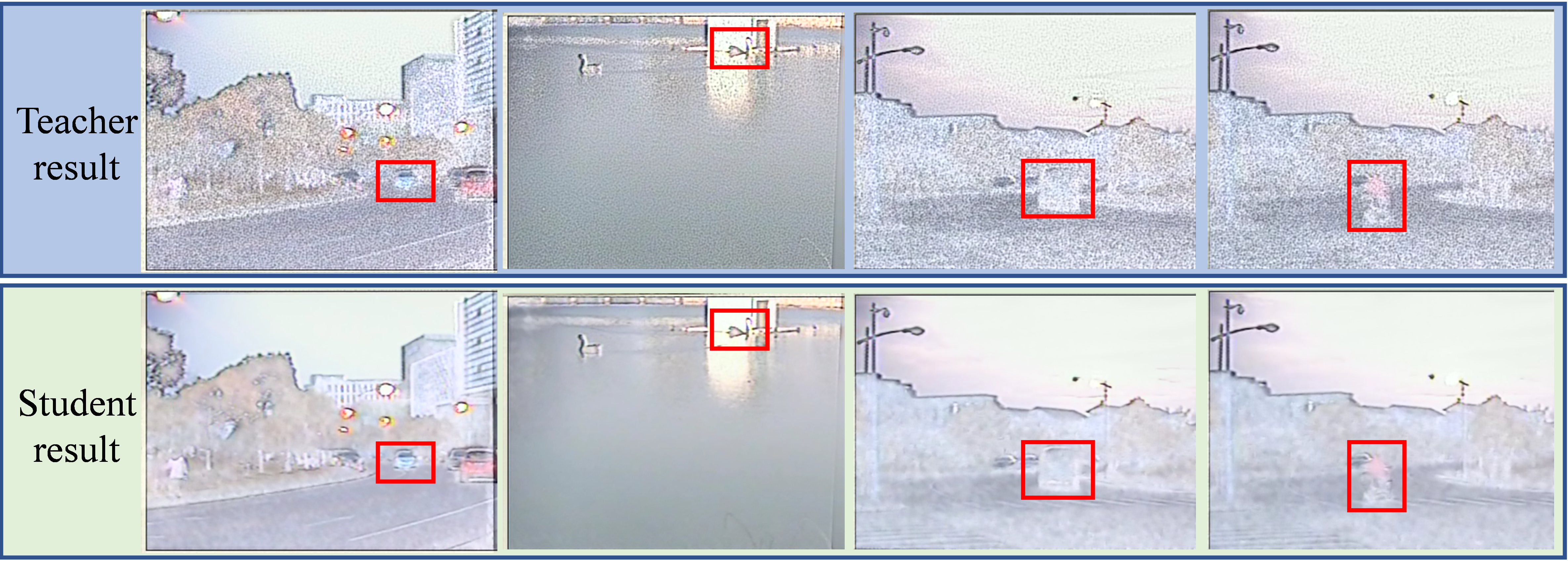}
    \caption{\textcolor{red}{Comparison results of the Teacher-Net and Student-Net. The objects are marked in red boxes.}}
    \label{fig11}
\end{figure}
\begin{figure}[htbp]
    \centering
    \includegraphics[width=1.0\linewidth]{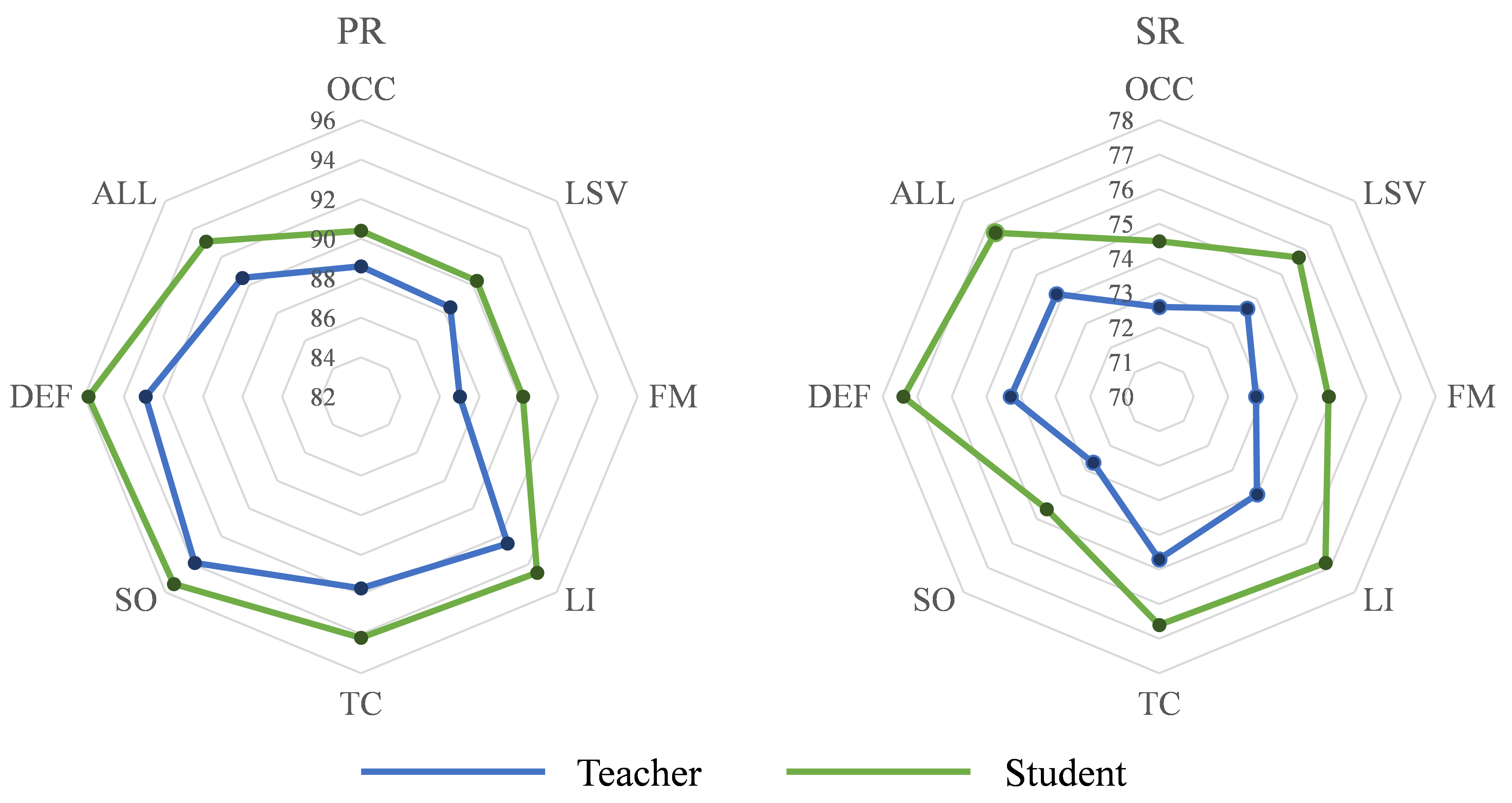}
    \caption{\textcolor{red}{Visualization of the tracking results on GTOT dataset.}}
    \label{fig12}
\end{figure}

\textcolor{red}{Figure 11 illustrates that the fusion modal outputs generated by the teacher network are subject to significant noise disturbances, whereas the student network produces results that successfully attenuate the impact of noise interference. In the outputs derived from the student network, the target features delineated by the red box are rendered with a greater level of detail in contrast to the outputs of the teacher network, thereby augmenting the tracker's ability for precise analytical discrimination and accurate target localization. Moreover, to substantiate the superiority of the proposed PGM, a comparative study was undertaken to assess the influence of fusion modal acquisition utilizing the teacher network versus the student network on the tracking results. The tracking efficacy is presented in the subsequent Fig. 12. Across the board, the PGM derived from the distillation architecture yields superior tracking performance, particularly in environments posing attributive challenges, thereby exemplifying its enhanced robustness and discriminative capabilities.}

To validate the feasibility of each individual contribution, four variants are implemented and tested on the GTOT and RGBT234 datasets. Specifically, X-Net-v1 serves as the base model, which simply fuses RGB and thermal features through element-wise addition. X-Net-v2 incorporates the PGM into the baseline network. X-Net-v3 combines both the PGM and FIM into the tracker based on the baseline. X-Net, the final version, is equipped with all proposed contributions. The ablation results for these variants are presented in Table \ref{table7}, from which the following conclusions are drawn: 1) Each proposed improvement significantly enhances the performance of the tracker. 2) By effectively leveraging the fused features from dual modalities, the PGM greatly improves the tracking precision and success rate of the baseline network. 3) The FIM effectively utilizes cues between modality space features to enhance the perception of modality, thereby improving the accuracy of localization. 4) The proposed DRM not only prevents misalignment but also contributes to achieving optimal tracking results.

\begin{table}[h]
\caption{Comparison results of the proposed method against the state-of-the-art trackers. attribute-based and overall performance are evaluated by PR scores ($\%$) and are produced on GTOT and RGBT234. The best results are marked in red.}
\begin{tabular}{cccccc}
\hline
Variants & PGM & FIM & DRM & GTOT      & RGBT234   \\ \hline
X-Net-v1  &     &     &     & 74.5/63.1 & 71.4/50.0 \\
X-Net-v2  & \checked   &     &     & 83.6/68.7 & 74.5/54.7 \\
X-Net-v3  & \checked   & \checked   &     & 90.4/74.5 & 78.8/57.6 \\
X-Net  & \checked   & \checked   & \checked   & 93.1/76.7 & 85.2/62.2 \\ \hline
\end{tabular}
\label{table7}
\end{table}

\section{Conclusion}
In this study, A high-performance X-Net is proposed, designed to effectively address the RGBT tracking task. The network integrates three carefully crafted and impactful design modules: PGM, FIM, and DRM, each contributing to a significant enhancement in tracking performance. The PGM integrates object cues from multi-modalities directly and mitigates noise interference effectively, operating as a plug-and-play pixel feature representation module through self-knowledge distillation learning. The FIM is introduced to address scale changes and facilitate cross-modal communication among multi-modal deep features, achieved through a combination of a spatial-dimensional shift strategy and a mixed feature interaction transformer. Finally, the DRM determines the re-tracking strategy using a refinement approach and optical flow algorithm. \textcolor{red}{The experimental results across three benchmark datasets confirm that the proposed X-Net outperforms existing state-of-the-art trackers and achieves performance gains of 0.47$\%$/1.2$\%$ in the average of PR/SR, respectively. However, the tracking performance and runtime efficiency of the proposed method on long sequence datasets with more challenging attributes need further enhancement.} As part of future work, our research will investigate the critical cues provided by multi-modal features that influence tracking performance and expand the knowledge distillation-based theoretical framework within the realm of RGBT tracking.

\section*{Acknowledgements}
This work is supported by the National Natural Science Foundation of China (Nos. 62266049). "Famous teacher of teaching" of Yunnan 10000 Talents Program. Key project of Basic Research Program of Yunnan Province (No. 202101AS070031). General project of National Natural Science Foundation of China (No. 81771928).
\section*{Data Availability Statement}
The data supporting the findings of this study are available at https://github.com/DZSYUNNAN/XNet.


\textcolor{red}{\bibliography{sn-bibliography}}

\end{document}